\documentclass[letterpaper]{article} 
\usepackage{aaai24}  
\usepackage{times}  
\usepackage{helvet}  
\usepackage{courier}  
\usepackage[hyphens]{url}  
\usepackage{graphicx} 
\usepackage{natbib}  
\usepackage{caption} 
\usepackage{algorithm}
\usepackage{algorithmic}

\usepackage{multirow}
\usepackage{enumitem}
\usepackage{newfloat}
\usepackage{listings}

\usepackage{xspace}
\usepackage{array}
\usepackage{ragged2e}
\usepackage{color}
\usepackage{bibentry}
\def\agc#1{\textcolor{black}{#1}} 
\def\tony#1{\textcolor{black}{#1}} 
\def\fj#1{\textcolor{black}{#1}} 
\def\lfj#1{\textcolor{black}{#1}} 

\DeclareCaptionStyle{ruled}{labelfont=normalfont,labelsep=colon,strut=off} 
\floatstyle{ruled}
\newfloat{listing}{tb}{lst}{}
\floatname{listing}{Listing}
\title{
Advancing Spatial Reasoning in Large Language Models: An In-Depth Evaluation and Enhancement Using the StepGame Benchmark
}

\author{
    Fangjun Li\textsuperscript{\rm 1},
    David C. Hogg\textsuperscript{\rm 1},
    Anthony G. Cohn\textsuperscript{\rm 1,2}
}
\affiliations{
    \textsuperscript{\rm 1}School of Computing, University of Leeds, Leeds, UK\\
    \textsuperscript{\rm 2}Alan Turing Institute, UK \\

    scfli@leeds.ac.uk, 
    d.c.hogg@leeds.ac.uk, 
    a.g.cohn@leeds.ac.uk
    
}

\begin{document}

\maketitle

\begin{abstract}
Artificial intelligence (AI) has made remarkable progress across various domains, with large language models like ChatGPT gaining substantial attention for their human-like text-generation capabilities. 
Despite these achievements, spatial reasoning remains a significant challenge for these models.  
Benchmarks like StepGame evaluate AI spatial reasoning, where ChatGPT has shown  unsatisfactory performance.  However, 
the presence of template errors in the benchmark has an impact on the evaluation results. 
Thus there is potential for ChatGPT to perform better if these template errors are addressed, leading to more accurate assessments of its spatial reasoning capabilities.
In this study, we refine the StepGame benchmark, providing a more accurate dataset for model evaluation. We analyze GPT's spatial reasoning performance on the rectified benchmark, identifying proficiency in mapping natural language text to spatial relations but limitations in multi-hop reasoning. 
We provide a flawless solution to the benchmark by combining template-to-relation mapping with logic-based reasoning. This combination demonstrates proficiency in performing qualitative reasoning on StepGame without encountering any errors.
We then address the limitations of GPT models in spatial reasoning. We deploy Chain-of-thought and Tree-of-thoughts prompting strategies, offering insights into GPT's ``cognitive process", and achieving remarkable improvements in \lfj{accuracy}. Our investigation not only sheds light on model deficiencies but also proposes enhancements, contributing to the advancement of AI with more robust spatial reasoning capabilities.

\end{abstract}

\section{Introduction}

Spatial reasoning, the ability to understand and navigate relationships in physical space, is a fundamental aspect of human cognition that significantly influences interactions with the environment. Enhancing spatial reasoning in AI models has the potential to enrich their comprehension of \agc{their} surroundings and response to user interactions, leading to more advanced and immersive user experiences \cite{alomari2022online}.
In recent years, AI has revolutionized numerous domains,
from healthcare to finance to entertainment. Notably, OpenAI's large language models (LLMs), such as ChatGPT and
GPT-4 \cite{OpenAI2023GPT4TR}, have gained significant attention for their human-like text generation capabilities. However, despite their impressive abilities, 
\agc{LLMs}
encounter challenges in 
\agc{many} 
logical reasoning aspects crucial for human communication, particularly spatial reasoning \fj{\cite{bang2023multitask, cohn2023dialectical}}.

\agc{One approach to evaluating spatial reasoning in an AI system is to use}
synthetic benchmarks 
\agc{such as}
StepGame \cite{shi2022stepgame} and SpartQA \cite{mirzaee2021spartqa}.
Unfortunately, models like ChatGPT have shown unsatisfactory performance on these benchmarks. Improving 
\agc{the} spatial reasoning capabilities \agc{of LLMs} remains a primary focus to enhance their overall performance and understanding of complex environments.

\agc{Whilst examining} the StepGame benchmark \agc{we discovered that it} contain\agc{s} template errors that distort model performance evaluations. These errors were previously overlooked, leading to studies conducted on a flawed benchmark, inaccurately assessing the capabilities of the LLMs \cite{bang2023multitask, yang2023coupling}. To rectify this issue, we present a more accurate version of the StepGame dataset for model evaluation, ensuring precise assessments of the models' true capabilities and limitations\footnote{The data associated with this paper are openly available from the University of Leeds Data Repository. \url{https://doi.org/10.5518/1468}. All code is available at \url{https://github.com/Fangjun-Li/SpatialLM-StepGame}.}.

\fj{We then conducted evaluation tests} on the rectified benchmark across various test subsets, few-shot sets, and models. We observed that larger GPT models 
demonstrate proficiency in mapping natural language text to spatial relations. However, they \fj{struggle with} multi-hop spatial reasoning.


Our goal is not merely to critique, but also to propose potential improvements. To this end, we provide a flawless solution to the benchmark, 
\fj{and}
explore different approaches to enhance the spatial reasoning ability of LLMs.

The solution we propose for the benchmark entails combining template-based sentence-to-relation mapping with logic-based spatial reasoning. The logical reasoner used in this approach comes from \cite{yang2023coupling}, where they integrated GPT-3 for the task.  GPT-3 was employed to parse spatial descriptions into symbolic spatial relation representations, which were then passed to the logical program for spatial reasoning.
This fusion resulted in significant improvement in StepGame, achieving state-of-the-art (SOTA) but not perfect results: around $90\%$ accuracy for lower hops and $88.3\%$ accuracy for 10-hop reasoning. They attributed $10.7\%$ faults to data-related issues. With our aforementioned work on rectifying the benchmark,  We take a step further to delve into the two components, analyzing the performance of each on our filtered version of the dataset. Remarkably, we achieved $100\%$ accuracy for almost all hops, with only 2 errors among 1000 test examples, which were due to GPT-3's incorrect semantic parsing. Building on this, we replaced the GPT-3 parser with our sentence-to-relation mapping method and combined it with the ASP reasoner, showcasing proficiency in performing qualitative reasoning without encountering any errors, 
thus demonstrating a method to achieve a perfect score on the corrected benchmark.

Neither our solution or the SOTA utilize LLMs for the actual spatial reasoning functionality.
Thus, we proceed to enhance GPT's capabilities as a \lfj{native spatial reasoner}. To achieve this, we employ Chain-of-Thoughts (CoT) and Tree-of-Thoughts (ToT) prompting strategies.

CoT \cite{wei2022chain} incorporates a sequence of intermediate reasoning steps to facilitate problem-solving.
However, when applied to StepGame, previous studies \cite{yang2023coupling} have shown that CoT does not consistently improve performance and may even reduce accuracy in complex $k$-hop reasoning tasks. This observation is attributed to the higher probability of errors occurring in lengthy CoT processes.
On the other hand, research on other tasks \cite{zhou2022least, creswell2022selection} has demonstrated that breaking down complex problems into simpler subproblems and solving them sequentially can be beneficial. Given the ambiguity in the decomposition of \agc{``}thoughts\agc{''\footnote{\agc{In this paper we use the word `thoughts' in the same way as is now being used in the literature on CoT and ToT, whilst noting that these are not thoughts in the human sense but rather generated coherent units of text , serving as intermediate steps in a problem solving setting, and without wishing to ascribe an anthropomorphic meaning to the word. }}} within CoT, we propose refining the CoT prompt to empower language models to perform better in spatial reasoning tasks.

On the other hand, \cite{yao2023tree} introduced ToT, a framework enabling LLMs to explore 
multiple reasoning paths, and they demonstrated its effectiveness in improving problem-solving capabilities across tasks like the game of 24, creative writing, and mini crosswords.
In our work, we customize the ToT approach for object-linking chain building, a crucial subproblem in addressing spatial reasoning benchmarks.

Our customized CoT method showcases its advantages more prominently in larger models such as GPT-4 and Davinci, maintaining accuracy even as the tasks become more complex.
Our ToT approach demonstrates its strengths on the three GPT models: on the largest model, GPT-4, we are able to maintain an accuracy of around 90\% even as the tasks become more complex. On Davinci, the accuracy is maintained at around 50\%, while Turbo achieves a lower level of accuracy at around 30\%. 

By identifying current deficiencies and proposing enhancements, we aim to contribute to the ongoing discourse in AI development, pushing the boundaries of what LLMs can achieve. Ultimately, our investigation can pave the way for the development of advanced, intuitive, and user-friendly AI systems with robust spatial reasoning capabilities.

\section{Related Work}

The field of spatial reasoning in language with artificial intelligence has evolved through sustained efforts over time, with significant advancements achieved through both traditional methods and modern LLMs.

Early strides in spatial reasoning \agc{in language} were marked by the development of formal structures to represent spatial relationships. \cite{kordjamshidi2010spatial} proposed \agc{a} spatial ontology to formalize the representation of spatial relations. This work laid the groundwork for the subsequent introduction of text-based spatial role labeling \cite{kordjamshidi2011spatial}, which aims to convert
text into formal spatial representations.


\lfj{Then comes synthetic tasks designed to evaluate the text understanding and spatial reasoning capabilities of learning algorithms. }
\fj{The positional reasoning task (Task 17) in the bAbI dataset \cite{weston2015towards} is for spatial reasoning and requires models to reason using one or two sentences, which makes this task comparatively simple. \cite{shi2022stepgame} advanced this field by creating the StepGame benchmark to evaluate multi-hop spatial reasoning in text, with richer variety in spatial relation descriptions.}
Both of these datasets emphasize directional spatial relations \lfj{\cite{cohn2001qualitative, skiadopoulos2001composing, cohn2008qualitative, chen2015survey}.}
Three spatial QA datasets: SpartQA\cite{mirzaee2021spartqa}, SPARTUN, and RESQ \cite{mirzaee2022transfer} expanded the resource landscape by encompassing wider-ranging spatial language expressions,
posing challenges for traditional logical programming, and are \lfj{important} benchmarks for evaluating LLMs' spatial reasoning capabilities.

Concurrently, the advent of LLMs such as OpenAI's ChatGPT \lfj{has} opened up fresh pathways for spatial reasoning. These models, leveraging transformer architectures, can generate human-like text and handle complex linguistic structures. 
However, while these models are indeed impressive, their capabilities in spatial reasoning are yet to be fully explored and exploited.
One recent approach to assess these capabilities was taken by \cite{bang2023multitask}, who put ChatGPT to the test using SpartQA and StepGame. Despite the generally advanced capabilities of ChatGPT, the model showed shortcomings in these tasks, signaling a need for further enhancements in the realm of spatial reasoning.

A promising technique known as `prompt engineering' \cite{bommasani2021opportunities} has been making its mark recently.
This approach involves crafting specific prompts to guide the responses of the models, leading to outputs that are more contextually apt and insightful. This method demonstrates significant potential in enhancing the capabilities of LLMs like ChatGPT in various domains \cite{li2022ontology}, including the challenging area of logical reasoning \cite{wang2023plan}.
For instance, when faced with multi-step reasoning tasks, a method called `few-shot chain-of-thought' (CoT) prompting \cite{zhang2022automatic} comes into play. 
These demonstrations enable LLMs to explicitly generate reasoning steps, thereby improving their accuracy in reasoning tasks.
This technique involves a handful of manually curated step-by-step reasoning demonstrations. 



As we review these developments, it is clear that while significant progress has been made, challenges remain in both traditional and LLM approaches to spatial reasoning. The limitations of models like ChatGPT indicate the need for continued research and enhancement strategies. This paper aims to contribute to this by examining these limitations more closely and proposing potential avenues for improvement. We aim to explore the limit of GPT as a general problem solver that explores its own thoughts and guides its own exploration with deliberate reasoning as heuristics.

\section{
The StepGame Benchmark for Evaluating Spatial Reasoning}
In this paper, we focus on StepGame, 
in line with other studies that evaluate ChatGPT's spatial reasoning proficiency using StepGame and SpartQA.
StepGame comprises story-question pairs in natural language, 
The objective is to answer questions regarding the spatial relations between two specified entities.
\lfj{The StepGame benchmark contains two sets of data: a \textit{clean} set where there are precisely $k$ facts given for any given $k$-hop instance, and a \textit{noise} set where there are more than $k$ facts given, and the extra facts are distracting.}

\subsection{Spatial Reasoning Types}

\begin{figure} [t]                
\centering
\includegraphics[width=0.48\textwidth]{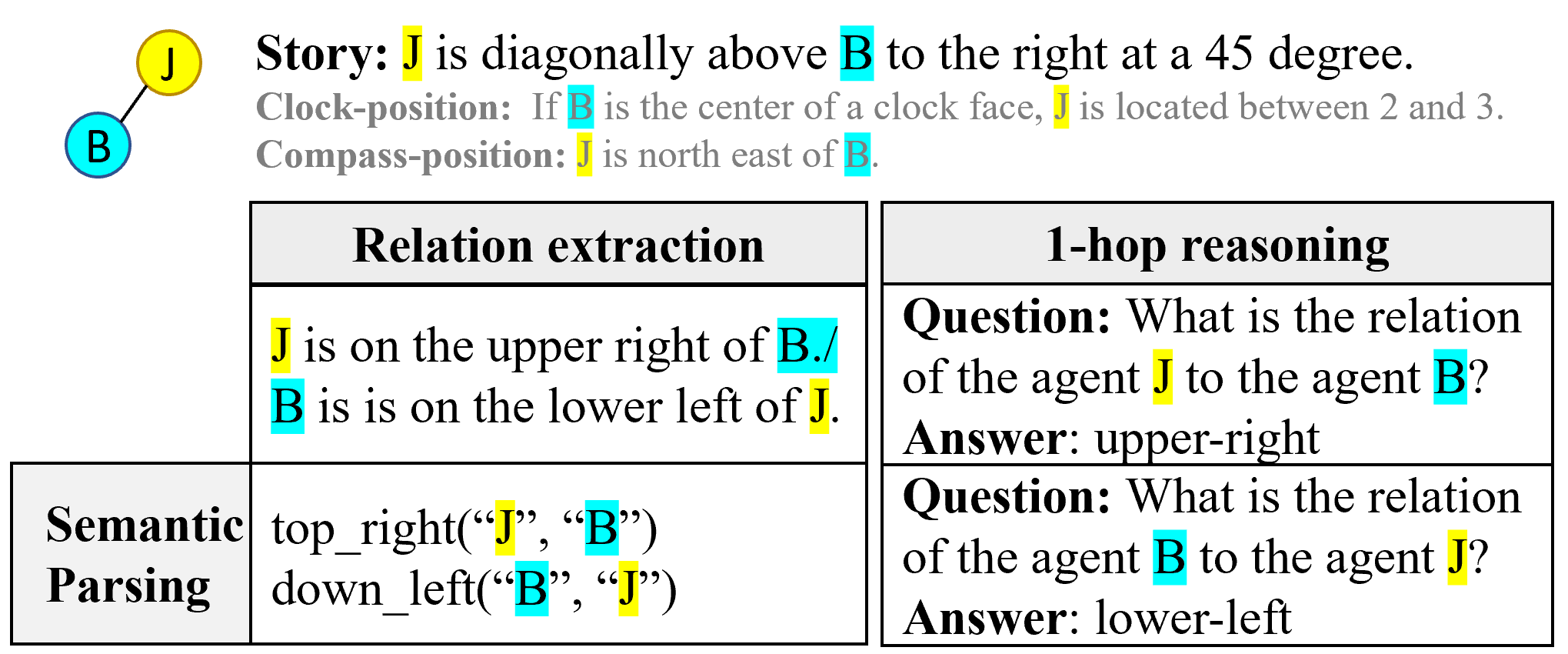}
\caption{An illustrative example for demonstrating relation extraction and 1-hop spatial  reasoning. }
\label{figure01}
\end{figure}

\begin{figure} [t]                
\centering
\includegraphics[width=0.47\textwidth]{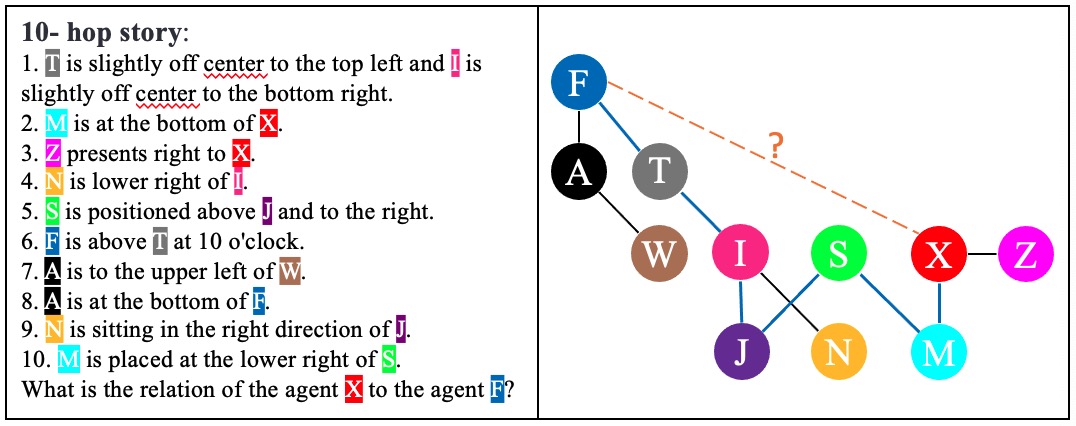}
\caption{Example of 10-hop reasoning, featuring a question regarding two entities that are not directly connected in the stories. \agc{The diagrams on the right do not form part of the input to the AI system but are for illustrative purposes only.} }
\label{figure02}
\end{figure}

\begin{itemize}[leftmargin=*]
\item  \textbf{1-Hop Spatial Reasoning}. In 1-hop reasoning, we are given a relation description between two entities and are asked about the spatial relation from one entity to the other. 
1-hop relation reasoning and relation extraction can be considered similar processes. 
As exemplified in Figure \ref{figure01}, consider
the story where \textit{J is diagonally above B to the right at a 45-degree angle}. The question \agc{is} \textit{What is the spatial relation of agent J to agent B?}. This is similar to relation extraction.  However, if we change the question to \textit{What is the spatial relation of agent B to agent J?}, it needs a reverse reasoning process \textit{top\_right(``J", ``B")} $\rightarrow$ \textit{down\_left(``B",``J")}. Both expressions are correct representations 
for relation extraction.

\item  \textbf{Multi-Hop Spatial Reasoning}.
Figure \ref{figure02} provides one example of 10-hop reasoning, which is from the `clean' set. 
The questions ask about the relation between two objects, either directly or indirectly connected. 
Multi-hop reasoning adds more complexity to the problem, as it involves a greater number of provided relations. To solve the problem, \lfj{one needs} to identify useful relations and then proceed with relation inference step by step.

\end{itemize}

\subsection{Problems with the Dataset}
\fj{Eight spatial relations (\textit{top, down, left, right, top-left, top-right, down-left, and down-right}) are utilized for the story generation of StepGame.}
These relations are expressed through sentences in natural language.
All sentences/statements are based on a crowd-sourced template base\footnote{\url{https://github.com/ZhengxiangShi/StepGame/blob/main/Code/template.py}}.
Each \lfj{``story"} is \fj{accompanied by a question that seeks to identify the relations between two objects, and it is labeled according to the intended relations at the time of story creation, rather than the actual sentences used.}
\fj{
A template is considered to contain an error if the meaning conveyed by the sentence does not align with the relationship that was intended to be expressed during the creation of stories and labels. 
}

Table \ref{table1} presents a detailed enumeration of errors in the relation-to-sentence mappings identified in StepGame. Out of the 214 templates examined, 14 were found to contain errors.
Of the eight different relationship mappings available, only $o_1\_above\_o_2$ and $o_1\_left\_o_2$ are devoid of mistakes. 
\agc{The question arises as to why there are so many errors in the crowd-sourced expressions; presumably this is down to insufficient quality control over the crowdworker reponses.}

\fj{
For each $k$ value, the StepGame dataset \lfj{includes} 10,000 test samples.}
Table \ref{table2} displays the percentage of examples containing sentences derived from incorrect templates, which hints at a rising trend in inaccuracies as \( k \) increases, suggesting a potential cumulative impact.




\newcolumntype{P}[1]{>{\centering\arraybackslash}p{#1}}
\begin{table}[t]
\small
\centering
\setlength{\tabcolsep}{0.9mm}
\begin{tabular}{l|p{6.7cm}}
\hline
\hline
\textbf{\small{Relation}} & \textbf{\small{Original Incorrect Template}}  \\
\hline
\multirow{4}{*}{{right}} & $o_2$ and $o_1$ are parallel, and $o_2$ on the right of $o_1$. \\
& $o_2$ and $o_1$ are parallel, and $o_2$ is to the right of $o_1$. \\
& $o_2$ and $o_1$ are horizontal and $o_2$ is to the right of $o_1$. \\
& \small{$o_2$ and $o_1$ are both there with the object $o_2$ is to the right of object $o_1$.}  \\
\hline 
\multirow{3}{*}{{below}} 
& $o_2$ is placed at the bottom of $o_1$. \\
& $o_2$ is at the bottom of $o_1$ and is on the same vertical plane. \\
& $o_2$ presents below $o_1$. \\
\hline
\multirow{2}{*}{{lowerleft}} & $o_2$ is there and $o_1$ is at the 10 position of a clock face. \\
& $o_2$ is positioned below $o_1$ and to the left. \\
\hline
\multirow{2}{*}{{upperright}} & Object A is above object $o_1$ and to the right of it, too.  \\
& $o_2$ is diagonally to the upper right of $o_1$. \\
\hline
{lowerright} & $o_1$ is to the right and above $o_2$ at an angle of about 45 degrees. \\
\hline
\multirow{2}{*}{{upperleft}} & $o_1$ is to the right and above $o_2$ at an angle of about 45 degrees. \\
& $o_1$ is diagonally left and above $o_1$. \\
\hline
\hline
\end{tabular}
\caption{
Incorrect sentence templates in StepGame. \fj{The Relation  column signifies relation for $o_1\_relation\_o_2$.}}
\label{table1}
\end{table}


\fj{Among these 14 incorrect templates, four cannot be remedied in existing StepGame benchmark examples.} 
\begin{itemize}[leftmargin=*]
    \item \textit{$o_1\_upperright\_o_2$}: Object A is above object $o_1$.
    \item \textit{$o_1\_upperleft\_o_2$ }: $o_1$ is diagonally left and above $o_1$.
    \item \textit{$o_1\_lowerright\_o_2\ |\ o_1\_upperleft\_o_2\ |\ o_1\_upperright\_o_2$}: $o_1$ is to the right and above $o_2$ at an angle of about 45 degrees.
    \item \textit{{$o_1\_lowerleft\_o_2\ |\ o_1\_upperleft\_o_2$}}: $o_2$ is there and $o_1$ is at the 10 position of a clock face.
\end{itemize}

\fj{The first and second templates are irreparable because it is impossible to identify what $o_2$ is when sentences are formed using them.}
\fj{
The third and fourth templates cannot be corrected since they were applied to multiple spatial relations, although each accurately represents just one. 
For example, \lfj{for} the sentence `Q is to the right and above P at an angle of about 45 degrees', three mapping relations exist: $Q\_upperright\_P$, $Q\_lowerright\_P$, and $Q\_upperleft\_P$. Although this sentence expresses the meaning $Q\_upperright\_P$, it is uncertain which candidate was used for the label.
For such templates, a unique correction could not be chosen, necessitating the removal of the sentences that use these template} \lfj{from the dataset}.

\begin{table}[t]
\centering
\small
\setlength{\tabcolsep}{0.2mm}
\begin{tabular}{l|c|c|c|c|c|c|c|c|c|c}
\hline

& k=1 & k=2 & k=3 & k=4 & k=5 & k=6 & k=7 & k=8 & k=9 & k=10 \\
\hline
 Clean  & 7.64 & 15.03 & 20.87 & 26.39 & 32.54 & 37.66 & 41.71 & 47.20 & 51.50 & \textbf{54.29} \\
 Noise   & 20.43 & 30.19 & 34.59 & 48.18 & 57.13 & 61.14 & 63.60 & 69.45 & 72.84 & \textbf{74.21} \\
\hline

\end{tabular}
\caption{Percentage of incorrect instances out of all instances over k=1--10 test sets.
} 
\label{table2}
\end{table}

\section{Methods}

\subsection{Solution for the Corrected Benchmark}

Our error-free approach is entirely logic-based, without the use of LLMs. We begin by performing template-based sentence-to-relation mapping, akin to semantic parsing. Then, we employ ASP for logical reasoning, utilizing the ASP reasoner introduced by \cite{yang2023coupling}. These two components operate independently:

\begin{itemize}[leftmargin=*]
    \item \textbf{Sentence-to-Relation Mapping}. 
    When presented with a natural language relation description $r$, we first identify the template used in $r$ through a comparison with the template base. This template is symbolized as ${o_0}\_\nu\_{o_1}$. Then, we convert this template form into a structured representation $\nu(o_0,o_1)$, where $o_0$ and $o_1$ correspond to the two objects mentioned in $r$, and $\nu$ signifies the spatial relation between $o_0$ and $o_1$. 
    Specifically, for questions inquiring about relations from the start object $o_0$ to the target object $o_t$, the template is $query\_o_0\_o_t$, and the corresponding ASP fact is represented as $query(o_0,o_t)$.
    Illustrative examples of this process can be 
    \agc{found} in Figure \ref{figure03}.
    
\begin{figure}    [t]  
\centering
\includegraphics[width=0.99\columnwidth]{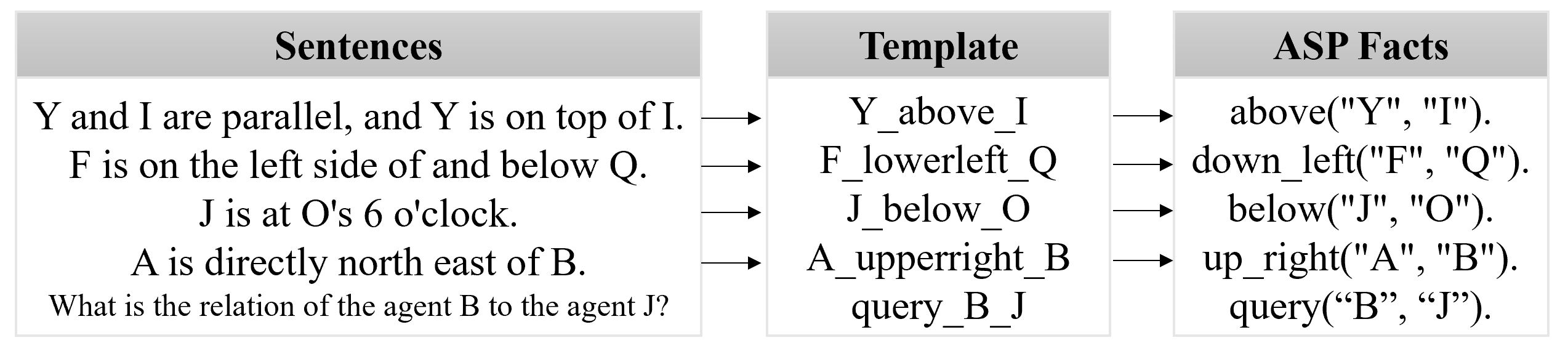}
\caption{Sentence-to-Relation Mapping Examples. 
}
\label{figure03}
\end{figure}
    \item \textbf{Logical Reasoning with ASP}. The logical facts $\nu(o_0,o_1)$, generated through semantic parsing for all relations in the story $R$, are used as input to the ASP module for spatial reasoning.  The ASP module was implemented using Clingo and includes rules specifically tailored for StepGame. These rules transform StepGame into a qualitative spatial reasoning problem in a 2D grid space. These rules incorporate offsets for 9 spatial relations, such as $\textit{offset}(right) = (0,1)$ and $\textit{offset}(lower\mbox{-}left) = (-1,-1)$. The main rule in the ASP module calculates the location of $o_0$ to $o_1$ by adding the offsets $\nu(o_0,o_1)$.
\end{itemize}

While this approach offers a solution to the StepGame benchmark challenge, it does require prior familiarity with the templates and mandates updates to the template base when confronted with new stories employing novel templates. 
In contrast, an LLM approach holds the potential to flexibly adjust to unfamiliar templates.
Additionally, the method's dependence on customized rules within the logical program constitutes another aspect to be mindful of.


\subsection{Chain-of-Thought (CoT) Prompting}

We devise\agc{d} a customized CoT for the spatial reasoning task.
The core idea of CoT is to introduce a chain of thoughts $c_1, \ldots, c_i, \ldots, c_n$ to bridge input $x$ and output $y$, where $i$ represents $i$-th step.
In our customized CoT for StepGame, $x$ consists of the task description, few-shot examples, relation story, and question, while $y$ represents the answer regarding the relations between the queried objects (from the start object $o_i$ to the target object $o_t$).
Each thought $c_i$ is to identify direct spatial connections between objects ($o_i$ and $o_{i+1}$). 
We take 
\agc{CoT}
a step further by decomposing each step of thought $c_i$ to explore the potential advantages of incorporating a coherent and detailed reasoning process.

\textbf{Thought Categorisation}.
We categorise the thought into three types:
link establishment thoughts $c^{link}$, relation mapping thoughts $c^{map}$, and coordinate calculation thoughts $c^{calcu}$.  At each reasoning step, these three \lfj{types} of thought are sequentially sampled as a continuous language sequence $c_i = [c^{link}_i, c^{map}_i, c^{calcu}_i]$ using the LLM. 

\begin{enumerate}[leftmargin=*]
\item \textbf{$c_i^{link}$}: 
Guide the LLM to examine all relations in the story ($R=[r^1,\ldots,r^j,\ldots,r^k]$) and select $r^j$ for the $i$-th step for $k$-hop reasoning, ensuring it directly describes the relation with $o_i$ and has not been used in any previous step.
For the start object ($i = 0$), we use the prompt ``Start with $o_0$. According to" and for the middle objects ($i \geq 1$), we use the prompt ``Then search for $o_i$. According to".
Full details of the prompts can be found in the Appendix\footnote{The ArXiv version of this paper includes the Appendix containing prompting examples.}.

\item \textbf{$c_i^{map}$}: 
Map $r^j$ to a simple relation description such as ``$o_i$ is to the $\nu$ of $o_{i+1}$," where $\nu$ represents the key spatial relation from $o_i$ to $o_{i+1}$. The prompt ``This means" helps the LLM perform this mapping.

\item \textbf{$c_i^{calcu}$}: 
Use $r^j$ to calculate the coordinates of $o_{i+1}$. We set $o_o$ at (0,0), and each spatial relation is assigned an offset to determine the positions of the objects. The prompt ``$o_{i+1}=o_i+\textit{offset}(r^j)= (x_{o_i}, y_{o_i})+(x_{\nu}, y_{\nu})=(x_{o_{i+1}}, y_{o_{i+1}})$" instructs the LLM on the calculation process. It computes the coordinates of $o_{i+1}$ and generates the output like ``Therefore, B is at $(x_{o_{i+1}}, y_{o_{i+1}})$."

\end{enumerate}


\subsection{Tree-of-Thoughts (ToT) Prompting}


Algorithm \ref{alg:ToT} is \lfj{designed} to enhance the reasoning chain-building process, allowing LLMs to consider different pathways. 
\agc{This is useful because} during the search for relations with an object, \lfj{distracting connections}
may arise, as shown in Figure \ref{figure02}. However, it is essential to follow 
\agc{a} correct sequence to successfully reach the target object. 
If 
\tony{an LLM} mistakenly tracks 
\tony{an incorrect} 
sequence, it could get stuck in a dead 
\agc{end} 
leading to incorrect reasoning conclusions such as ``The story does not provide direct spatial information."



The algorithm \lfj{initiates by prompting the LLM to set up the initial tree state, denoted as $S_0$, using the input $x$, which comprises a story and a question. $S_0$ is in the form ``chain: $o_0$ \texttt{->}, target: $o_t$, unused: $R$". $R$ represents all connections between objects in the story, in the form of $object1$-$object2$.}
Then it proceeds to construct
\lfj{a}
linking chain from $o_0$ to $o_t$ in iterative steps, wherein for the $i$-th step $(1\leq i \leq 10)$, the LLM considers the tree state $S_{i-1}$ built up to that step.
\lfj{If no state $s$ in $S_{i-1}$ reaches $o_t$, the LLM
\tony{is prompted} to generate $j$ candidate thoughts 
\tony{for each $s$ in the current set of states, $S_i$}
\tony{($j=2$ in this paper)}. 
$G$ prompts the LLM to search for a potential object $o_{i}$ connected
to the current object $o_{i-1}$ from the unused relations $R_{i-1}^{unused}$. 
A check is made ($CheckExtn(c))$) to see if the proposal made is a real candidate extension.
For all candidate thoughts, $V$ prompts the LLM to evaluate the state to determine if the chain can proceed with $o_i$ and the updated $R_{i-1}^{unused}$ to reach $o_t$. The top-rated $b$ tree states \tony{in} $S_i^{'}$ 
\tony{are} selected as $S_i$.
When there is a state $s_f$ \tony{which} reaches $o_t$, the L will be prompted with the linking chain construction prompt (Appendix D.4) to form the final links $l$.
}


\begin{itemize} [leftmargin=*]
\item \textbf{Thought Generation $G(s, j)$}.  
Given a tree state $s$, we 
let the LLM propose \tony{$j$} thoughts using the thought generation prompt
``Use relations listed in unused relations to enumerate all potential expansions of the chain by considering unused relations that exhibit a direct link to the last object within the chain.''
\lfj{In our experiment, we set $j=2$, meaning that we instruct the LLM to generate content twice for each 
state 
$s \in S_{i-1}$. 
}  

\begin{algorithm} [t]
\caption{Our ToT Approach}
\label{alg:ToT}
\textbf{Require}: LLM, input $x$
\begin{algorithmic}[1] 
\STATE \( S_0 \leftarrow Init(x) \) \\
\STATE $i \leftarrow  1$
\WHILE{no $s_f \in S_{i-1}$ has arrived at \( o_t \)}
    \STATE $S_i^{'} \leftarrow  \{s \cdot c  | c \in G(s,j) \wedge ChainExtn(c) \wedge s \in S_{i-1} \}$
    \STATE   \textbf{if} $S_i^{'} = \emptyset$  \textbf{then return} failure

    \STATE $S_i \leftarrow select(b,\{\langle s, y\rangle| s\in S'_i \wedge y = \Sigma_1^n \sigma(V(s)) \})$
    \STATE $i = i+1$
\ENDWHILE
\STATE \textbf{return} $Link(s_f)$
\end{algorithmic}
\end{algorithm}

\item \textbf{State Evaluation $V(s)$}.     
Our approach involves a classification methodology, using the designed value prompt
``Evaluate whether the chain can reach the target (sure/likely/impossible). If the chain has already reached the target, it's `sure'. If the unused relations include the current object, it's `likely'. If there are no unused relations that include the current object, it's `impossible'.'' 
This prompt  guides the LLM to sequentially examine all 
newly generated states
$s \in S_i^{'}$ \lfj{$n$ times -- using the stochasticity of the LLM with a non zero temperature to increase the reliability of the scoring. }
\lfj{The three types of outputs - `sure', `likely', and `impossible' - 
\tony{are} converted into \tony{numerical} scores using a function $\sigma( )$ to facilitate the selection process among all \tony{newly generated states}.}

\item \textbf{Search Algorithm}
The choice between utilizing breadth-first search (BFS) or depth-first search (DFS) depends on the tree structure. In the StepGame benchmark, the tree depth is limited ($depth \leq 10$), and the number of thought candidates $k$ for each step is also limited ($width \leq 3$ in most cases). However, a deeper search does not necessarily guarantee better results. In certain scenarios,  $o_0$ and $o_t$ may be directly connected in one relation statement, allowing for shorter linking chains between them, which is preferable. Therefore, we opt for BFS to maintain all promising states. 
We set the breadth width $b=3$, maintaining the three most promising linking-chain states per step. 
The criterion for stopping searching is set when the linking chain arrives at the target object.

\end{itemize}

Our ToT approach \lfj{is used} to construct the reasoning chain from $o_0$ to $o_t$. Subsequently, the spatial relation between these objects is computed
following the previous CoT prompting method, with the use of $c^{map}$ and $c^{calcu}$.

\section{Experimental Design}
\subsection{Model Settings}
We use \agc{the} Azure OpenAI Service for ChatGPT (3.5-Turbo) and GPT-3 (Davinci), and GPT-4 API access.  To yield more concentrated and deterministic results, we set the temperature to 0 in CoT experiments.
\lfj{In ToT experiments, we follow \cite{yao2023tree}, setting the temperature to} 0.7 for generating varied thought proposals.
The remaining parameters were left at the standard configurations for these models.



\subsubsection{Different Test Subsets}

It is common practice in the studies cited \cite{bang2023multitask}, \cite{yang2023coupling} to use a subset of 30 or 100 test examples from the full set of 10,000 for each \( k \) value. While this method helps in conserving token usage, it could potentially introduce biases or 
inaccurate estimations of the model performance.

We examine the effect of the number of test examples. Specifically, we wanted to determine whether evaluating on a limited number of test examples could introduce inaccuracies. To achieve this, we conducted tests on a clean, filtered test set for \( k \)-hop reasoning ($k \in [1,10]$), thereby covering a range of task complexities. 
Tests were carried out on 30, 100, and 1000 test examples to assess the impact of the number of test examples on the evaluation. 

\subsubsection{Different Few-Shot Sets}
We created three different few-shot prompting sets to evaluate the influence of input examples in prompts.
\begin{itemize}[leftmargin=*]
    \item \textit{clean 5shot(1,3,5,7,10)}: Create a prompt consisting of five examples, with one example each from tasks requiring 1-hop, 3-hop, 5-hop, 7-hop, and 10-hop reasoning.
    \item \textit{clean 10shot}: Formulate a prompt using ten examples, each one derived from a distinct \( k \)-hop task in clean set.
    \item \textit{clean 5shot separate}: Construct a prompt for each \( k \)-hop reasoning task, utilizing five examples from the corresponding \( k \)-hop training set as few-shot examples.
\end{itemize}

\begin{figure*}  
\centering
\includegraphics[width=1.97\columnwidth]{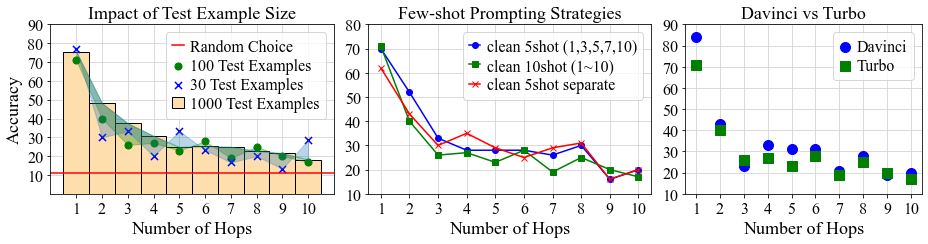}
\caption{
Accuracy comparison for varying numbers of hops (1-10) \lfj{on the clean test set}. On the left, we show the performance variation of  
\fj{the Turbo model with \textit{10shot} prompting} 
over different test set sizes (30, 100, and 1000 examples). The middle section illustrates the performance of the Turbo model under three distinct prompting settings: \textit{5shot(1,3,5,7,10)}, \textit{10shot}, and \textit{5shot separate}. The right portion showcases the performance of two models - Davinci and Turbo - using \textit{10shot} prompting.
}
\label{figure04}
\end{figure*}

\section{Experimental Results}



\subsection{Evaluation Results}

\subsubsection{Influence of Scale of Test Examples}

We employ the \textit{clean 10shot} prompting setting. The results are presented in the left subplot of  Figure \ref{figure04}.
Upon evaluation of the expanded test set comprising 1000 examples, the model shows a uniform decrement in performance as $k$ increases from 1 to 10.
This trend indicates the increased complexity 
as the number of hops increases.
With smaller test sets of 100 or 30 examples, the trend is less consistent, and there are occasional increases in performance at certain hop levels. The variance in performance, particularly for the 30-example test set, may indeed be larger. This could be due to the smaller sample size providing less comprehensive coverage of the potential range of tasks, leading to more fluctuations in performance.
This indicates larger test sets can provide a more stable and reliable indicator of a model's performance across different complexity levels (i.e., number of hops).


\subsubsection{Influence of Prompting Examples}

The middle  subplot in figure \ref{figure04} indicates that the choice of prompting strategy can impact the model's ability to handle tasks of varying complexity.
Similar to the previous data, all prompting strategies show a trend of decreasing accuracy as the number of hops increases. This trend is consistent and suggests that the complexity of the tasks grows with the number of hops.

The performances of the three methods are close. While differences exist at specific hop levels, no single method consistently outperforms the others across all hop levels.
Interestingly,  clean 5shot (1,3,5,7,10) performs better than clean 10shot (1$\sim$10) at almost every hop level. This suggests that selecting examples from a wider range of hop levels (1,3,5,7,10) can be more beneficial than having an example from each hop level from 1 to 10.

\subsubsection{Influence of Models}

As indicated in a recent study \cite{ye2023comprehensive}, Turbo demonstrates comparable performance to Davinci across many tasks. However, it falls short in the machine reading comprehension, part-of-speech, and relation extraction tasks, potentially owing to its smaller model size.
The StepGame spatial reasoning task requires the comprehension of sequential spatial connections and the ability to draw deductions from them.
According to the right subplot of Figure \ref{figure04}, the Davinci model generally outperforms the Turbo model across varying levels of task complexity (number of hops).  The differences in performance between the two models are more significant at lower complexity levels, but they appear to converge as the complexity increases.

\subsection{Results of the Improved Methods}

\subsubsection{Resolution for the Benchmark}
The results of our resolution (sentence-to-relation mapping + ASP-based reasoning) are displayed in the `Map+ASP' row of Table \ref{table7}. The numbers in the table indicate \lfj{accuracy} scores, with higher values indicating better performance. This demonstrates the proficiency achieved in spatial relation mapping and multi-hop spatial reasoning, all without encountering any errors.

\subsubsection{GPT for Relation Extraction + ASP for Reasoning} 

We analyze the performance of GPT in the relation extraction subtask, as outlined in Table \ref{table5}. Curie has the highest number of wrong predictions across different relations, Davinci and Turbo show better performance.

\begin{table}
    \centering
    \setlength{\tabcolsep}{0.4mm}
    \begin{tabular}{p{2.5cm}|p{0.8cm}|p{1cm}|p{1.7cm}|p{1.7cm}}
       \hline
       \hline
        &  left/ right & above /below & lower\_left/ upper\_right &  lower\_right/ upper\_left \\ 
       \hline
       total & 44 & 53 & 50 & 53\\
       \hline
       text-curie-001 & 11 & 41 & 30 & 37\\ 
       \hline
       text-davinci-003 & \textbf{0} & \textbf{0} & \textbf{0} & 2\\ 
       \hline
       gpt-3.5-turbo & 2 & 2 &3  & \textbf{1} \\
       \hline
       \hline
    \end{tabular}
    \caption{The relation extraction performance of GPT. The numbers in rows 2-4 are incorrect predictions \lfj{numbers}.}
    \label{table5}
\end{table}

The state-of-the-art results achieve\agc{d} by \cite{yang2023coupling} (using GPT\lfj{-3} for semantic parsing and ASP for reasoning) are presented in the ``SOTA" row of Table \ref{table7}. They achieve approximately 90\% accuracy for lower hops and 88.3\% accuracy for 10-hop reasoning. They attribute 10.7\% of the inaccuracies to data-related concerns.

We provide an evaluation of their approach onthe corrected dataset, with the results displayed in the ``Curie+ASP" and ``Davinci+ASP" rows. Among the 1000 test examples (100 for each k), only 2 errors were encountered with Davinci.
\lfj{caused by}
semantic parsing: the sentence ``If E is the center of a clock face, H is located between 2 and 3." was parsed incorrectly as $right(``H", ``E")$,
\lfj{but supposed to be $up\_right(``H", ``E")$}.




\subsubsection{CoT and ToT}

The experimental results in Table \ref{table7} involving GPT-4 and ToT are based on a test set comprising 20 instances considering token usage,
while for Davinci and Turbo, we used a larger test set of 100 samples.
\fj{The results for the base and CoT methods were obtained using the \textit{5shot separate} prompting} \lfj{on the \textit{clean} set}. \lfj{All the ToT\_CoT results presented in the table involve the use of GPT-4 for building the linking chain, followed by the application of Turbo, Davinci, and GPT-4 for CoT reasoning with the constructed linking chain.}
The GPT-4 model exhibits superior performance across nearly all settings.
With the basic input-output prompt, despite starting at 100\% accuracy for $k=1$, its accuracy dips to 25\% for $k=10$, indicating that even the most powerful GPT model struggles to maintain accuracy as task complexity rises. \lfj{Humans would probably find this challenging too.}

\begin{table} 
    \small
    \setlength{\tabcolsep}{0.2mm}
    \begin{tabular}{c|c|c|c|c|c|c|c|c|c|c|c}
    \hline
    \hline
    \multicolumn{2}{c|}{}&\textbf{k=1}  &\textbf{k=2} &\textbf{k=3}  &\textbf{k=4} &\textbf{k=5}  &\textbf{k=6} & \textbf{k=7} &\textbf{k=8} &\textbf{k=9}  &\textbf{k=10} \\
    \hline
    \multicolumn{2}{c|}{Map+ASP} &  \textbf{100}  &  \textbf{100}  &  \textbf{100}  &  \textbf{100}  &  \textbf{100}  &  \textbf{100}  &  \textbf{100}  &  \textbf{100}  &  \textbf{100}  &  \textbf{100}  \\
    \hline
    \hline
    \multicolumn{2}{c|}{Curie+ASP}& 46& 43 &42 &59 &67 &67 &57 &56 &58 &61 \\  
    \multicolumn{2}{c|}{Davinci+ASP} &  \textbf{100}  &  \textbf{100}  &  99  &  \textbf{100}  &  \textbf{100}  &  99  &  \textbf{100}  &  \textbf{100}  &  \textbf{100}  &  \textbf{100}  \\
    \multicolumn{2}{c|}{SOTA} &  92.6  &  89.9   &  89.1   &  93.8   &  92.9  &   91.6  &  91.2  &  90.4  &   89.0  &   88.3  \\
    \hline
    \hline
    \multirow{3}{*}{Turbo} & base & 62 & 43 & 30 & 35 & 29 & 25 & 29 & 31 & 16 & 20 \\
    & CoT & / &34 &  40 & 36	& 28 & 28  &	26  	&  31 & 25 & 24   \\
     &\footnotesize{ToT\_CoT} & / & /& 35 & 35 & 25 & 45 & 15 & 40 & 40 & 35 \\
    \hline
    \multirow{3}{*}{Davinci}  & base &77 & 42 & 21	& 26 & 25& 30& 23& 23& 22& 22 \\
     &CoT & / & 48 & 53 & 46 & 46 & 48 & 40 & 45 & 41 & 32    \\
     &\small{ToT\_CoT}  & / & /& 65 & 50 & 45 & 60 & 50 & 50 & 55 & 50 \\
    \hline
    \multirow{3}{*}{GPT-4}  &base &100 &70 & 55 & 45 & 40 & 25 & 40 & 35 & 35 & 25\\
    & CoT & /  & \textbf{80} &	75 &  \textbf{95}	& 85	& 85 & \textbf{90}  &	 80 &  60 &  65\\
    &\small{ToT\_CoT}  & / & /& \textbf{85} & 85 & \textbf{90} & \textbf{90}  & 85 & \textbf{90} & \textbf{100}  & \textbf{95} \\
    \hline
    \hline
    \end{tabular}
    \caption{Accuracy comparison of  GPT models on \lfj{revised} StepGame using different methods. 
    }
    \label{table7}
\end{table}


With the implementation of our  CoT and ToT approach, the GPT-4 model demonstrates significant performance enhancements for more complex tasks (ranging from $k=2$ to $k=10$). 
Our ToT and CoT method considerably enhances the performance of the Davinci and GPT-4, particularly in larger hops.
For the Turbo model, although our CoT method brings improvements as $k$ increases, the gains are not as profound as those observed with the Davinci and GPT-4.
This could be attributed to the long length of our prompts, requiring a nuanced understanding of coordinates and relations. 

\section{Conclusion}



\fj{
This \lfj{paper} has introduced a \lfj{revised version} of the StepGame benchmark, correcting template errors that distort model performance evaluations, leading to a more accurate evaluation of the spatial reasoning capabilities of \lfj{AI systems attempting the challenge}. We highlight Davinci and Turbo's abilities in mapping texts to spatial relations and their limitations in multi-hop spatial reasoning.
Our solution combines template-to-relation mapping with logic-based reasoning, effectively addressing challenges \lfj{in this task}. We also enhance LLMs' spatial reasoning ability through prompt engineering, using CoT and ToT strategies.
}

\fj{
This paper focuses on StepGame; future studies could extend our findings to other benchmarks. Our methods are suitable for adaptation to various 2D grid-based directional spatial tasks, such as the bAbI (task 17). This adaptation would involve customizing the template for the ASP-based solution and modifying task descriptions and few-shot examples for CoT and ToT approaches. For tasks that require a combination of directional, topological, and distance reasoning, like SpartQA, it would be necessary to integrate additional rules and ontology into both the ASP program and the prompts to LLMs for effective solution development.
}


\fj{The effective resolution of the StepGame benchmark prompts a need for more challenging versions.}
\lfj{While having a well-defined set of spatial relations converted into natural language using a set of templates is appealing, it leads to controlled natural language which is more amenable to special purpose reasoning. Finding a way to generate more naturalistic problem statements automatically would therefore  
be highly desirable.}
\fj{Additionally, the current independent use of LLMs and logic programs suggests a potential research direction towards integrating these tools for more comprehensive and cohesive problem-solving strategies.}



\section{Acknowledgments}

This work has been partially supported by Microsoft Research - Accelerating Foundation Models Research program, with the provision of Azure resources to access GPT. This work was also partially supported by the Turing’s Defence and Security programme through a partnership with the UK government in accordance with the framework agreement between GCHQ and The Alan Turing Institute.

\section{Author Contributions} AGC and DCH proposed the initial line of work. FL designed the actual implementation, performed all the evaluations, and wrote the initial paper draft. DCH and AGC supervised FL. All authors contributed to subsequent paper revisions.

\bibliography{aaai24}

\clearpage
\appendix

\textbf{A. Example Prompts for Base}

\texttt{\footnotesize{Given a story about spatial relations among objects, answer the relation between two queried objects. Possible relations are: overlap, above, below, left, right, upper-left, upper-right, lower-left, and lower-right. If a sentence in the story is describing clock-wise information, then 12 denotes above, 1 and 2 denote upper-right, 3 denotes right, 4 and 5 denote lower-right, 6 denotes below, 7 and 8 denote lower-left, 9 denote left, 10 and 11 denote upper-left. If the sentence is describing cardinal directions, then north denotes above, east denotes right, south denotes below, and west denotes left.\\
\\
Story:\\
Q is to the right of O and is on the same horizontal plane.\\
Q is slightly off center to the top left and M is slightly off center to the bottom right.\\
X and E are next to each other with X on the top and E at the bottom.\\
O is sitting at the upper right position to E.\\
W is on the right side and below M.\\
What is the relation of the agent W to the agent E?\\
Answer: lower-right\\
\\
$\cdots$ \\
\\
Story:\\
1. The object E is positioned directly above the object W.\\
2. E is sitting at the upper right position to I.\\
3. W is placed at the upper left of C.\\
4. L is over there and Y is on the left.\\
5. C and Y are both there with the object Y below the object C.\\
6. What is the relation of the agent E to the agent Y?\\
}}

\textbf{B. Example Prompts for CoT in \cite{yang2023coupling}}

\texttt{\footnotesize{Given a story about spatial relations among objects, answer the relation between two queried objects. Possible relations are: overlap, above, below, left, right, upper-left, upper-right, lower-left, and lower-right. If a sentence in the story is describing clock-wise information, then 12 denotes above, 1 and 2 denote upper-right, 3 denotes right, 4 and 5 denote lower-right, 6 denotes below, 7 and 8 denote lower-left, 9 denote left, 10 and 11 denote upper-left. If the sentence is describing cardinal directions, then north denotes above, east denotes right, south denotes below, and west denotes left.\\
\\
Story:\\
Q is to the right of O and is on the same horizontal plane.\\
Q is slightly off center to the top left and M is slightly off center to the bottom right.\\
X and E are next to each other with X on the top and E at the bottom.\\
O is sitting at the upper right position to E.\\
W is on the right side and below M.\\
What is the relation of the agent W to the agent E?\\
Answer: We first link W and E using the relations in the story. W is to the lower-right of M. M is to the lower-right of Q. Q is to the right of O. O is to the upper-right of E. So the answer is lower-right.
\\
$\cdots$ \\
\\
Story:\\
1. The object E is positioned directly above the object W.\\
2. E is sitting at the upper right position to I.\\
3. W is placed at the upper left of C.\\
4. L is over there and Y is on the left.\\
5. C and Y are both there with the object Y below the object C.\\
6. What is the relation of the agent E to the agent Y?\\
}}

\textbf{C. Example Prompts for Our CoT}

\texttt{\footnotesize{Given a story about spatial relations among objects, answer the relation between two queried objects. Possible relations are: overlap, above, below, left, right, upper-left, upper-right, lower-left, and lower-right. If a sentence in the story is describing clock-wise information, then 12 denotes above, 1 and 2 denote upper-right, 3 denotes right, 4 and 5 denote lower-right, 6 denotes below, 7 and 8 denote lower-left, 9 denote left, 10 and 11 denote upper-left. If the sentence is describing cardinal directions, then north denotes above, east denotes right, south denotes below, and west denotes left. In all the spatial relations, assume that all agents occupy a position on a grid point of equally spaced points in the vertical and horizontal directions and that agents occupy the nearest grid point consistent with the spatial relation. The offsets of 9 spacial relations: offset(overlap) = (0,0); offset(above) = (0,1); offset(below) = (0,-1); offset(left) = (-1,0); offset(right) = (1,0); offset(upper-left) = (-1,1); offset(upper-right) = (1,1); offset(lower-left) = (-1,-1); offset(lower-right) = (1,-1).\\
\\
Story:\\
1. Q is to the right of O and is on the same horizontal plane.\\
2. Q is slightly off center to the top left and M is slightly off center to the bottom right.\\
3. X and E are next to each other with X on the top and E at the bottom.\\
4. O is sitting at the upper right position to E.\\
5. W is on the right side and below M.\\
What is the relation of the agent W to the agent E?\\
Reasoning:\\
Let's suppose W is at (0,0). We can connect W and E using the relations given in the story.\\
Start with W. According to 5, "W is on the right side and below M." This means M is to the upper-left of W. M= W+ offset(upper-left) = (0,0)+(-1,1)=(-1,1). Therefore, M is at (-1,1).\\
Then search for M. According to 2, "Q is slightly off center to the top left and M is slightly off center to the bottom right." This means Q is to the upper-left of M. Q= M+ offset(upper-left) = (-1,1)+(-1,1)=(-2,2). Therefore, Q is at (-2,2).\\
Then search for Q. According to 1, "Q is to the right of O and is on the same horizontal plane." This means O is to the left of Q. O= Q+ offset(left) = (-2,2)+(-1,0)=(-3,2). Therefore, O is at (-3,2).\\
Then search for O. According to 4, "O is sitting at the upper right position to E." This means E is to the lower-left of O. E= O+ offset(lower-left) = (-3,2)+(-1,-1)=(-4,1). Therefore, E is at (-4,1).\\
We've reached E. So, considering W(0,0) and E(-4,1), W is to the lower-right of E.\\
Answer: lower-right\\
\\
$\cdots$ \\
\\
Story:\\
1. The object E is positioned directly above the object W.\\
2. E is sitting at the upper right position to I.\\
3. W is placed at the upper left of C.\\
4. L is over there and Y is on the left.\\
5. C and Y are both there with the object Y below the object C.\\
6. What is the relation of the agent E to the agent Y?\\
}}

\textbf{D. Example Prompts for Our ToT}

\textbf{D.1. Tree state initialization prompt}

\texttt{\footnotesize{
Provided with a sequence of statements that define the spatial relationships among various objects, your task is to detail the subsequent actions. This includes initiating the chain of connections, identifying the target object, and enumerating all links between objects from the statements.\\
\\
Input: 1. Q is to the right of O and is on the same horizontal plane. 2. Q is slightly off center to the top left and M is slightly off center to the bottom right. 3. X and E are next to each other with X on the top and E at the bottom. 4. O is sitting at the upper right position to E. 5. W is on the right side and below M. What is the relation of the agent W to the agent E?\\
Possible next steps:\\
chain: W -$>$, target: E, unused: 1. Q-O, 2. Q-M, 3. X-E, 4. O-E, 5. W-M.\\
\\
$\cdots$ \\
\\
Input: \{input\}\\
Possible next steps:\\
}}

\textbf{D.2. Thought generation prompt}

\texttt{\footnotesize{
Use relations listed in unused relations to enumerate all potential expansions of the chain by considering unused relations that exhibit a direct link to the last object within the chain.\\
\\
Input: chain: G -$>$, target: Q, unused: 1. C-R, 2. L-Q, 3. C-J, 4. J-E, 5. T-A, 6. G-N, 7. G-A, 8. L-Y, 9. R-Q, 10. Y-T.\\
Possible next steps:\\
The last object within the chain is G, and the unused relations 6. G-N and 7. G-A include G. 
relation  chain: G -$>$ N (use 6) -$>$, target: Q, unused: 1. C-R, 2. L-Q, 3. C-J, 4. J-E, 5. T-A, 7. G-A, 8. L-Y, 9. R-Q, 10. Y-T. \\
chain: G -$>$ A (use 7) -$>$, target: Q, unused: 1. C-R, 2. L-Q, 3. C-J, 4. J-E, 5. T-A, 6. G-N, 8. L-Y, 9. R-Q, 10. Y-T.\\
\\
$\cdots$ \\
\\
Input: \{input\}\\
Possible next steps:\\
}}

\textbf{D.3. State evaluation prompt}

\texttt{\footnotesize{
Evaluate whether the chain can reach the target (sure/ likely/impossible). If the chain has already reached the target, it's 'sure'. If the unused relations include the current object, it's 'likely'. If there are no unused relations that include the current object, it's 'impossible'.\\
\\
chain: F ->, target: X, unused: 1. Y-F, 2. X-Y, 3. I-Q, 4. A-Q, 5. N-W, 6. N-A, 7. F-O, 8. O-W.
The current object is F, there are unused relations that include F (1. Y-F, 7. F-O).\\
likely\\
\\
chain: L -> Q (use 2) ->, target: Q, unused: 1. C-R, 3. C-J, 4. J-E, 7. G-A, 8. L-Y, 9. R-Q.\\
The chain already reaches the target object Q.\\
sure\\
\\
chain: G -> N (use 6) ->, target: Q, unused: 1. C-R, 2. L-Q, 3. C-J, 4. J-E, 5. T-A, 8. L-Y, 9. R-Q, 10. Y-T.\\
The current object is N, and there are no unused relations that include N.\\
impossible\\
\\
\{input\}\\
}}



\textbf{D.4. Linking chain construction prompt}

\texttt{\footnotesize{
Given an input about spatial relations among objects, build the linking chain between the two queried objets.\\
\\
Input:\\
1. H is above S with a small gap between them. 2. S is positioned below I. 3. P is on the top side to I. What is the relation of the agent S to the agent P?\\
Steps:\\
chain: S -$>$, target: P, unused: 1. H-S, 2. S-I, 3. P-I.\\
chain: S -$>$ I (use 2) -$>$, target: P, unused: 1. H-S, 3. P-I.\\
chain: I -$>$ P (use 3) -$>$, target: P, unused: 1. H-S.\\
Answer: S -$>$ I (use 2) -$>$ P (use 3)\\
\\
$\cdots$ \\
\\
Input:\\
\{input\}\\
}}

\textbf{D.5. Spatial relation reasoning prompt}

\texttt{\footnotesize{
Given a story about spatial relations among objects, answer the relation between two queried objects. Possible relations are: overlap, above, below, left, right, upper-left, upper-right, lower-left, and lower-right. If a sentence in the story is describing clock-wise information, then 12 denotes above, 1 and 2 denote upper-right, 3 denotes right, 4 and 5 denote lower-right, 6 denotes below, 7 and 8 denote lower-left, 9 denote left, 10 and 11 denote upper-left. If the sentence is describing cardinal directions, then north denotes above, east denotes right, south denotes below, and west denotes left. In all the spatial relations, assume that all agents occupy a position on a grid point of equally spaced points in the vertical and horizontal directions and that agents occupy the nearest grid point consistent with the spatial relation. The offsets of 9 spacial relations: offset(overlap) = (0,0); offset(above) = (0,1); offset(below) = (0,-1); offset(left) = (-1,0); offset(right) = (1,0); offset(upper-left) = (-1,1); offset(upper-right) = (1,1); offset(lower-left) = (-1,-1); offset(lower-right) = (1,-1).\\
Story:\\
1. Q is to the right of O and is on the same horizontal plane.\\
2. Q is slightly off center to the top left and M is slightly off center to the bottom right.\\
3. X and E are next to each other with X on the top and E at the bottom.\\
4. O is sitting at the upper right position to E.\\
5. W is on the right side and below M.\\
What is the relation of the agent W to the agent E?\\
Linking chain: W -$>$ M (use 5) -$>$ Q (use 2) -$>$ O (use 1) -$>$ E (use 4)\\
Reasoning:\\
Let's suppose W is at (0,0). We can analyze the relation of W to E by following the linking chain and considering the relations provided in the story step by step.\\
Start with W. According to 5, "W is on the right side and below M." This means M is to the upper-left of W. M= W+ offset(upper-left) = (0,0)+(-1,1)=(-1,1). Therefore, M is at (-1,1).\\
Then come to M. According to 2, "Q is slightly off center to the top left and M is slightly off center to the bottom right." This means Q is to the upper-left of M. Q= M+ offset(upper-left) = (-1,1)+(-1,1)=(-2,2). Therefore, Q is at (-2,2).\\
Then come to Q. According to 1, "Q is to the right of O and is on the same horizontal plane." This means O is to the left of Q. O= Q+ offset(left) = (-2,2)+(-1,0)=(-3,2). Therefore, O is at (-3,2).\\
Then come to O. According to 4, "O is sitting at the upper right position to E." This means E is to the lower-left of O. E= O+ offset(lower-left) = (-3,2)+(-1,-1)=(-4,1). Therefore, E is at (-4,1).\\
We've reached E. So, considering W(0,0) and E(-4,1), W is to the lower-right of E.\\
Answer: lower-right\\
\\
$\cdots$ \\
\\
Story:\\
\{input\}\\
Linking chain: \{chain\}\\
}}

\end{document}